\documentclass[11pt]{article}

\usepackage[final]{acl}

\usepackage{times}
\usepackage{latexsym}
\usepackage{layouts}
\usepackage[T1]{fontenc}

\usepackage[utf8]{inputenc}

\usepackage{microtype}

\usepackage{inconsolata}

\usepackage{graphicx}
\usepackage{hyperref}
\usepackage{url}

\usepackage{xspace}

\usepackage{amsthm,amsmath,amssymb,amsfonts}
\usepackage{wrapfig}
\usepackage{caption}
\usepackage{subcaption}
\usepackage{makecell} 
\usepackage{booktabs}       
\usepackage{xcolor,colortbl}
\usepackage[most]{tcolorbox}
\usepackage{tabularx}
\usepackage{multirow}
\usepackage{arydshln}
\usepackage{stackengine}  
\usepackage{soul} 
\usepackage{listings}
\usepackage{fvextra}

\colorlet{lightblue}{blue!15}
\colorlet{lightorange}{orange!15}
\colorlet{lightgreen}{green!15}
\colorlet{lightgray}{gray!15}
\colorlet{lightyellow}{yellow!15}
\colorlet{lightpink}{pink!30}

\newcommand{\escolor}[1]{{\colorbox{lightblue}{#1}}}
\newcommand{\hscolor}[1]{{\colorbox{lightgreen}{#1}}}
\newcommand{\phdcolor}[1]{{\colorbox{lightorange}{#1}}}

\newcommand{\es}{\escolor{Elementary School}\xspace}
\newcommand{\esshort}{\escolor{Elementary}\xspace}
\newcommand{\hs}{\hscolor{High School}\xspace}
\newcommand{\hsshort}{\hscolor{High School}\xspace}
\newcommand{\phd}{\phdcolor{Graduate School}\xspace}
\newcommand{\phdshort}{\phdcolor{Graduate}\xspace}

\newcommand{\lowcomplexity}{\escolor{Low}\xspace}
\newcommand{\mediumcomplexity}{\hscolor{Medium}\xspace}
\newcommand{\highcomplexity}{\phdcolor{High}\xspace}

\newcommand{\model}[1]{\texttt{#1}\xspace}

\newcommand{\scienceqa}{\textsc{ScienceQA}\xspace}
\newcommand{\eliwhygpt}{\textsc{ELI-Why (GPT-4)}\xspace}
\newcommand{\eliwhyhuman}{\textsc{ELI-Why (Human)}\xspace}
\newcommand{\augustetal}{\textsc{Scientific Papers}\xspace}
\newcommand{\cleardataset}{\textsc{CLEAR}\xspace}
\newcommand{\ie}{\textit{i.e.},\xspace}
\newcommand{\eg}{\textit{e.g.},\xspace}
\newcommand{\n}{\textbackslash n}

\newcommand{\readme}{\textsc{ReadMe++}\xspace}
\newcommand{\readmeavg}{\textsc{ReadMe++ (Avg)}\xspace}
\newcommand{\readmemax}{\textsc{ReadMe++ (Max)}\xspace}
\newcommand{\metaraterreadability}{\textsc{Meta Rater (readability)}\xspace}
\newcommand{\metaraterprofessionalism}{\textsc{Meta Rater (professionalism)}\xspace}
\newcommand{\llmzeroshot}{\textsc{LLM-as-a-judge (0-shot)}\xspace}
\newcommand{\llmfiveshot}{\textsc{LLM-as-a-judge (5-shot)}\xspace}
\newcommand{\llmshotcontinuous}{\textsc{LLM-as-a-judge (continuous 0-100)}\xspace}

\newif\ifcomments
\commentstrue
\ifcomments
    \providecommand{\kat}[2][]{{\protect\color{teal}{(Kat: #1 #2)}}}
    \providecommand{\parker}[2][]{\textcolor{purple}{(Parker: #1 #2)}}
    \providecommand{\anoop}[2][]{\textcolor{orange}{(Anoop: #1 #2)}}
\else
    \providecommand{\kat}[1][]{}
    \providecommand{\parker}[1][]{}
    \providecommand{\anoop}[1][]{}
\fi

\newcommand{\postspace}{\vskip -3mm}
\newcommand{\minipostspace}{\vskip -1.5mm}

\newcommand*{\Scale}[2][4]{\scalebox{#1}{$#2$}}%
\title{
Readability Reconsidered: \\ A Cross-Dataset Analysis of Reference-Free Metrics
}

\author{
 \textbf{Catarina G. Belem\textsuperscript{1}},
 \textbf{Parker Glenn\textsuperscript{2}},
 \textbf{Alfy Samuel\textsuperscript{2}},
 \textbf{Anoop Kumar\textsuperscript{2}},
 \textbf{Daben Liu\textsuperscript{2}},
\\
\\
 \textsuperscript{1}University of California Irvine,
 \textsuperscript{2}Capital One
\\
 \small{
   \textbf{Correspondence:} \href{mailto:cbelem@uci.edu}{cbelem@uci.edu}
 }
}

\begin{document}
\maketitle
\begin{abstract}
Automatic readability assessment plays a key role in ensuring effective and accessible written communication.
Despite significant progress, the field is hindered by inconsistent definitions of readability and measurements that rely on surface-level text properties.
In this work, we investigate the factors shaping human perceptions of readability through the analysis of 897 judgments, finding that, beyond surface-level cues, information content and topic strongly shape text comprehensibility.
Furthermore, we evaluate 15 popular readability metrics across five English datasets, contrasting them with six more nuanced, model-based metrics.
Our results show that four model-based metrics consistently place among the top four in rank correlations with human judgments, while the best performing traditional metric achieves an average rank of 8.6.
These findings highlight a mismatch between current readability metrics and human perceptions, pointing to model-based approaches as a more promising direction.
\end{abstract}

%
%
%
%
\section{Introduction}
\label{ssec:introduction}
\textit{Readability assessment} can be used to determine the level of comprehension of a piece of text~\citep{DuBay2004ThePO,CollinsThompson2014ComputationalAO}. 
In domains such as science communication~\citep{Kerwer2021,august-et-al-2023-paper-plain}, 
health~\citep{Friedman2006,Hershenhouse2024},
law~\citep{Curtotti-legal-example,Cheong-et-al-2024-legal-motivation}, 
and education~\citep{vajjala-lucic-2018-onestopenglish}, 
readability assessment plays a key role in making information accessible to individuals regardless of their background or cognitive needs~\citep{CollinsThompson2014ComputationalAO}. 
It is important for highly-specialized fields characterized by dense jargon and complex language~\citep{Friedman2006,han2024use},
as well as for applications engaging with users of varied familiarity with the domain~\citep{joshi2025eliwhyevaluatingpedagogicalutility,puech-etal-2025-towards}.

One challenge in advancing automatic readability assessment is that \textit{readability} is an overloaded term, measured in different ways by prior work. 
Some studies treat readability as \textit{text difficulty}, using  surface-level properties such as word length, word frequency, and various word type counts~\citep{Flesch1948ANR,Kincaid1975DerivationON,leroy-et-al-2008-eval-online-beyond-read-formulas}.
Others broaden the definition of readability to consider syntactic and discourse-level organization, including cohesion and coherence properties~\citep{Graesser2004,Petersen2007,pitler-nenkova-2008-revisiting,feng-etal-2010-comparison,Eslami2014,zhuang-etal-2025-meta}.
A third line of work views readability as a combination of text characteristics and information content~\citep{xia-etal-2016-text,august-et-al-2024}. 

Taken together, the diversity of interpretations highlight the difficulty of pinning down readability, and have led to the continued use of proxy metrics that may not fit the task, domain, or are misaligned with human comprehension judgments~\citep{ahmed-2023-beyond,liu-lee-2023-hybrid,han2024use}.

\section{Related Work}
\label{ssec:relatedwork}

\paragraph{Readability Datasets. } 
Despite growing interest in readability assessment, high-quality datasets remain scarce~\citep{xia-etal-2016-text}. 
Existing document-level datasets can be subdivided into \textit{parallel} corpora~\citep{vajjala-lucic-2018-onestopenglish,august-et-al-2024,joshi2025eliwhyevaluatingpedagogicalutility} and non-parallel corpora~\citep{lu2022learn-scienceqa,crossley-et-al-2023-CLEAR-dataset} and span various tasks and content type, including literary and informational~\citep{crossley-et-al-2023-CLEAR-dataset}, academic~\citep{august-et-al-2024}, or information-seeking content~\citep{lu2022learn-scienceqa,joshi2025eliwhyevaluatingpedagogicalutility}.
Recently, sentence-level datasets have also been introduced~\citep{arase-etal-2022-cefr,naous-etal-2024-readme}.

\paragraph{Readability Metrics.}
While human judgments remain the gold standard for readability evaluation, their collection is often time-consuming and expensive~\citep{rooein-etal-2024-beyond}.
Automated metrics have emerged as a cheaper and quicker alternative.
Examples include metrics relying on basic linguistic features, including sentences, words, and syllables counts, average reading time~\citep{metric-avg-reading-time-demberg-keller-2008}, language model perplexity~\citep{CollinsThompson2014ComputationalAO,pitler-nenkova-2008-revisiting}, and fraction of functional~\citep{leroy-et-al-2008-eval-online-beyond-read-formulas,Leroy-et-al-2010-funct-words-ratio-in-health} or uncommon words~\citep{august-et-al-2024}. 
Surface-form features have been further combined to form   \textit{readability tests}, such as the Automatic Reading Index~\citep{senter1967automated}, Dale-Chall Readability Score~\citep{dalle-chall-readability}, Flesch-Kincaid Reading Ease~\citep{Flesch1948ANR}, and Linsear Write Formula~\citep{Klare1974-linsearwrite}. 
Despite critiques of brittleness~\citep{rooein-etal-2024-beyond,CollinsThompson2014ComputationalAO} and limited domain suitability~\citep{Leroy-et-al-2010-funct-words-ratio-in-health}, these formulas continue to be used.
Recently, both fine-tuning~\citep{arase-etal-2022-cefr,naous-etal-2024-readme} and LLM-as-a-judge approaches~\citep{rooein-etal-2024-beyond,trott-riviere-2024-measuring} have been proposed to capture more abstract and nuanced aspects of readability. However, since these methods rely on implicitly learned representations, they are regarded as less interpretable than those grounded in surface-level textual features.

\section{How Do Humans Perceive Readability?}
\label{sec:human-perceptions-readability}
Given the divergent definitions of readability and continued reliance on surface-form metrics, we take a human-centric perspective, asking: \textit{What guides human perceptions of readability?} 
To address this question, we analyze a subset of the \eliwhygpt~\citep{joshi2025eliwhyevaluatingpedagogicalutility} dataset, designed to study whether LLMs can generate explanations tailored to various readability levels. The dataset comprises GPT-4–generated explanations for 299 ``Why'' questions, each annotated by humans into three readability levels—\esshort, \hsshort, and \phdshort—along with accompanying rationales justifying their judgments.
Each question–explanation pair was independently rated by three annotators, and final labels were determined via majority vote. 
For additional details, see the original paper. 
Table~\ref{tab:qualitative-examples:human-rationales} (in Appendix) shows randomly selected examples of human rationales for each readability level. 

\paragraph{Exploring Human Rationales.}
Although \citet{joshi2025eliwhyevaluatingpedagogicalutility} collected human rationales supporting readability judgments, their analysis primarily focuses on the labels themselves, offering limited insight into the factors shaping human perceptions. We complement their study by providing a quantitative perspective on the key factors driving human text comprehension through the analysis of human rationales. 
Two authors of this paper annotated the human-provided readability rationales for 90 ELI-Why question–answer pairs, balanced evenly across classes. Building on the original human annotation instructions, each rationale was labeled with one or more of the following categories:
\begin{itemize}
\item \textit{Wording/Terminology}: presence of scientific words, abbreviations, or complex synonyms;
\item \textit{Sentence Structure}: comments on sentence length or the number of concepts;
\item \textit{Examples/Analogies}: mentions of examples or analogies as key factors;
\item \textit{Details and Depth}: mentions of the presence or absence of details;
\item \textit{Curriculum-based}: links the information content or topic to a specific education level.
\end{itemize}
\begin{figure}[tb]
  \includegraphics[width=0.95\linewidth]{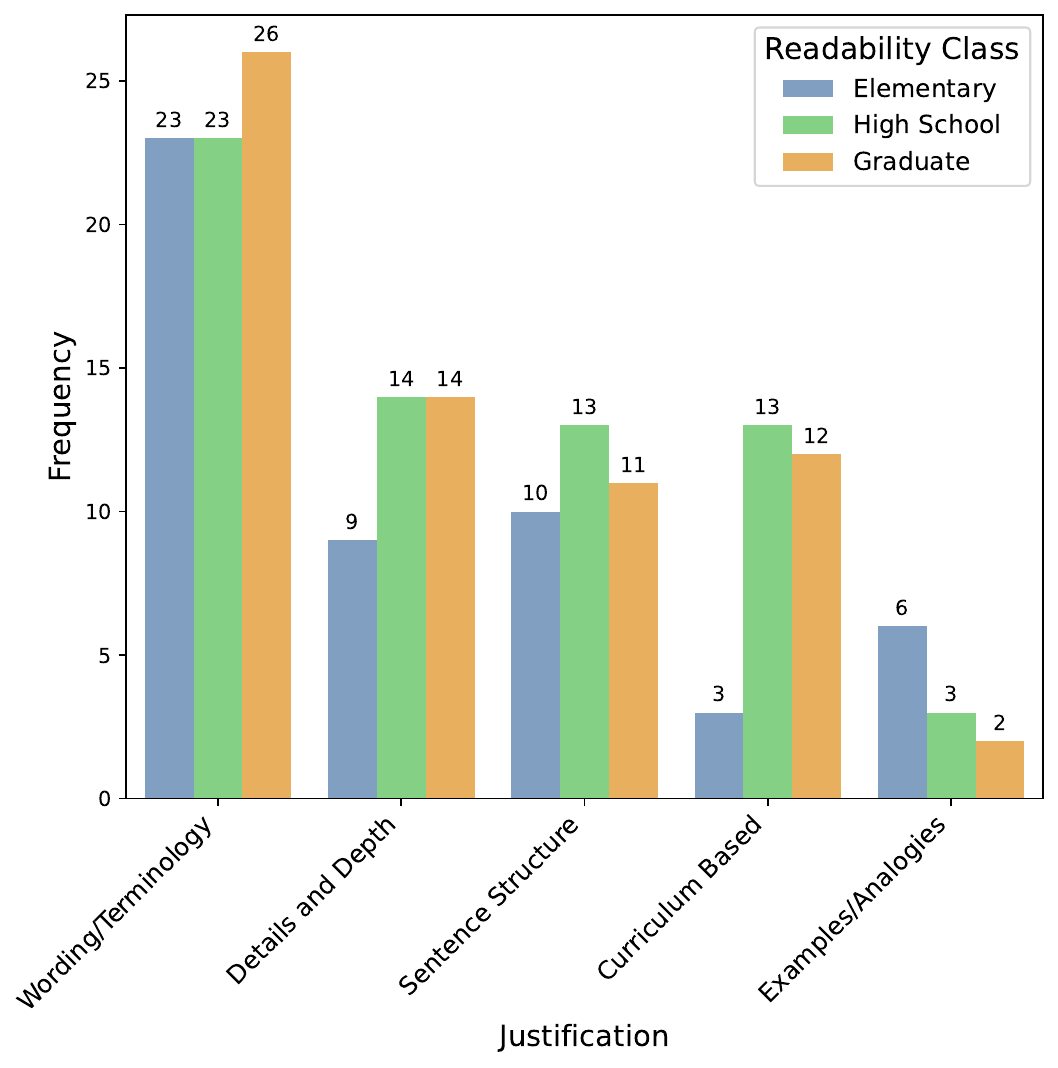} 
  \caption{\textbf{Distribution of justification reasons across 90 examples in \eliwhygpt}. Counts are based on the consensus over 2-way annotations.}
  \label{fig:readability_annotations}
  \postspace\minipostspace
\end{figure}

Figure \ref{fig:readability_annotations} shows the consensus vote across readability classes. The average sample-level Jaccard index for the obtained annotations is 0.91, indicating high agreement between the two annotators. 
\textit{Wording/Terminology} emerges as the predominant rationale for readability judgments, with annotators distinctions in lexical complexity (\eg ``Words like adherence are too advanced for elementary school'') or simplicity (``uses basic words''). 
The \textit{Curriculum-based} category is invoked far more often to justify \hsshort and \phdshort judgments than \esshort, with annotators noting that ``The scientific terms... require an introductory background or some foundational knowledge'' or that ``a concept that will be brought up in chemistry classes in undergrad.'' Conversely, \textit{Examples/Analogies} is disproportionately used to support \esshort judgments, with comments such as ``Examples are... what you'd say to a toddler'' or ``The analogies used... make it more accessible to elementary level.''. Notably, both categories rely on comprehension and common-sense reasoning that go beyond surface-level textual properties.

\section{Re-evaluating Readability Metrics}
\label{sec:evaluating-readability}
Motivated by the gap between surface-form textual cues and human perceptions of readability, we investigate how well existing readability metrics correlate with human judgments across five diverse datasets (see statistics in Table \ref{tab:datasets-statistics}).\footnote{Additional dataset details are available in Appendix \ref{appdx:dataset:statistics}.}

\subsection{Datasets}
\label{ssec:eval-readability-datasets}
\begin{table*}[tb]
  \small
  \centering
  \begin{tabular}{p{3.5cm} cc p{5cm} cc}
    \toprule
    \thead{\textbf{Dataset}}
        & \thead{\textbf{Size}}
        & \thead{\textbf{Label} \textbf{Type}}
        & \thead{\textbf{Labels} }
        & \thead{\textbf{Avg.} \\ \textbf{\textsc{\#Words}}}  
        & \thead{\textbf{Avg.} \\ \textbf{\textsc{\#Sents}}} \\
    \midrule
    \augustetal{\scriptsize~\citep{august-et-al-2024}}
        & 180 & categorical & \scriptsize \lowcomplexity $\prec$ \mediumcomplexity $\prec$ \highcomplexity  & 65.93 & 2.22 \\
    \addlinespace
    \cleardataset{\scriptsize~\citep{crossley-et-al-2023-CLEAR-dataset}}
        & 1000 & continuous & N/A & 199.23 & 9.45 \\
    \addlinespace
    \eliwhygpt{\scriptsize~\citep{joshi2025eliwhyevaluatingpedagogicalutility}}
        & 897 & categorical & \scriptsize \esshort $\prec$ \hsshort$\prec$ \phdshort  & 144.21 & 6.97 \\
    \addlinespace
    \eliwhyhuman{\scriptsize~\citep{joshi2025eliwhyevaluatingpedagogicalutility}}
        & 117 & categorical & \scriptsize \esshort $\prec$ \hsshort$\prec$ \phdshort & 99.03 & 4.22 \\
    \scienceqa{\scriptsize~\citep{lu2022learn-scienceqa}}
    & 2295 & categorical & \scriptsize Grade 1 $\prec$ Grade 2 $\prec$ ... $\prec$ Grade 12  & 183.08 & 13.26 \\
    \bottomrule
  \end{tabular}
  \caption{Dataset statistics, including dataset size, readability label type (continuous vs categorical), average number of words and sentences across examples.}
  \label{tab:datasets-statistics}
  \postspace\minipostspace
\end{table*}

\textbf{\augustetal}~\citep{august-et-al-2024} consists of 180 total query-focused summaries about 10 different academic papers (\eg What did the paper do?)
and cover topics from public policy to nanotechnology.
Summaries were carefully curated by an expert science writer to reflect three levels of complexity: \lowcomplexity, \mediumcomplexity, and \highcomplexity. 

\textbf{\cleardataset}~\citep{crossley-et-al-2023-CLEAR-dataset} contains 4.7k text excerpts sourced from open digital libraries including Project Gutenberg and Wikipedia. The texts are self-contained and cover both cover literary and informational content. Approximately 111k pairwise readability judgments from 1.1k annotators were aggregated under a Bradley-Terry model to obtain continuous readability scores.

\textbf{\eliwhygpt}~\citep{joshi2025eliwhyevaluatingpedagogicalutility} includes 897 GPT-4–generated explanations tailored to three readability levels---\esshort, \hsshort, and \phdshort---each annotated with human-assigned labels and rationales. 
Likewise, \textbf{\eliwhyhuman} is a smaller dataset with 123 answers that were manually curated.

\textbf{\scienceqa}~\citep{lu2022learn-scienceqa} is a multi-modal science reasoning dataset consisting of 21k multiple choice questions sourced from K-12 curriculum, covering various subjects (\eg natural science, language science, and social science). 
Each example is associated with a reference solution (or \textit{explanation}) and reference knowledge (or \textit{lecture}), both of which are written at the readability level of the intended student audience. We randomly sample 200 text-only examples per grade for our analysis.

\subsection{Metrics}
\textbf{Surface-form metrics} consist of direct counts of properties of the text, such as characters, syllables, monosyllables, polysyllables, words, and sentences. These also include other specialized variants such as estimated reading time in seconds, number of difficult words, and functional words.

\textbf{Psycholinguistic metrics}, known as \textit{readability tests}, are typically formulated as weighted sums of ratios involving surface-form properties.
For instance, \textit{Automatic Readability Index} is based on characters-to-words and words-to-sentences ratio~\citep{senter1967automated}, 
the \textit{Flesch Kincaid Reading Ease} on words-to-sentences and syllables-to-words~\citep{Flesch1948ANR}, 
and \textit{Dale-Chall Readability} on the fraction of difficult words and words-to-sentences ratio~\citep{dalle-chall-readability}. 
An exception is the \textit{Linsear Write Formula}, which distinguishes easy from hard words using syllable counts and computes their frequencies in a text sample~\citep{Klare1974-linsearwrite}. 
We additionally report values for other popular metrics~\citep{coleman1975computer,gunning1952technique,smog-index}.\footnote{We refer readers to Appendix \ref{appdx:automated-metrics:long-definition} for additional details.}
\label{ssec:eval-readability-metrics}

\textbf{Model-based metrics} are categorized into two main classes: 
\textit{fine-tuned metrics}~\citep{zhuang-etal-2025-meta} 
and \textit{LLM-as-a-judge metrics}~\citep{llm-as-a-judge-neurips2023}.
In this work, we use two fine-tuned metrics based on ModernBERT \citep{warner2024smarter} -- \metaraterreadability and \metaraterprofessionalism, which were recently introduced to evaluate texts along readability and professionalism dimensions, respectively~\citep{zhuang-etal-2025-meta}.
The former considers factors such as clarity, coherence, vocabulary complexity, and sentence structure with the goal of assessing whether a reader can understand a written text, whereas the latter relies on the depth and content accessibility to determine the degree of expertise or knowledge required to comprehend a text.
Additionally, we include a complementary BERT-based metric---\readme~\citep{naous-etal-2024-readme}---which predicts readability in terms of language learning capabilities through the use of the 6-point Common European Framework of Reference for Languages scale.

We test three different LLM-as-a-judge approaches, including the zero-shot continuous score approach by~\citet{trott-riviere-2024-measuring} (dubbed \textsc{LLM-as-a-judge continuous 0-100}). We also test a categorical setting, in which a model is tasked with predicting one of three readability labels - \esshort, \hsshort or \phdshort. We prompt the model with the same instructions provided to human annotators in ~\citet{joshi2025eliwhyevaluatingpedagogicalutility} and, in the 5-shot setting, include the five example annotations (two \esshort, two \phdshort, one \hsshort).
All LLM-as-a-judge approaches are performed using \model{Llama-3.3-70B-Instruct} with greedy decoding (\texttt{temperature=0}).

\subsection{Results \& Discussion}
\label{ssec:results}

\begin{table*}[t]
\resizebox{\linewidth}{!}{
\begin{tabular}{ll ccccc r}
\toprule
\textbf{Type} & \textbf{Metric}
    & \thead{\textbf{\augustetal}\\\citep{august-et-al-2024}}  
    & \thead{\textbf{\cleardataset}\\\citep{crossley-et-al-2023-CLEAR-dataset}} 
    & \thead{\textbf{\eliwhygpt}\\\citep{joshi2025eliwhyevaluatingpedagogicalutility}} 
    & \thead{\textbf{\eliwhyhuman}\\\citep{joshi2025eliwhyevaluatingpedagogicalutility}} 
    & \thead{\textbf{\scienceqa}\\\citep{lu2022learn-scienceqa}} 
    & \thead{\textbf{Avg.} \\ \textbf{Rank}}\\
\midrule
\multirow{9}{*}{Surface-form}
    & \# Words & 0.16$^*$ & -0.06$^*$ & 0.46$^*$ & 0.15 & 0.28$^*$ & 17.0 \\
    & \# Sentences & 0.25$^*$ & 0.23$^*$ & 0.38$^*$ & -0.07 & 0.09$^*$ & 17.0 \\
    & Avg. Sentence Length & -0.15 & -0.25$^*$ & 0.21$^*$ & 0.40$^*$ & 0.39$^*$ & 16.4 \\
    & Avg. Reading Time (s) & 0.20$^*$ & -0.23$^*$ & 0.47$^*$ & 0.25$^*$ & 0.32$^*$ & 14.8 \\
    & \# Syllables & 0.22$^*$ & -0.28$^*$ & 0.47$^*$ & 0.28$^*$ & 0.33$^*$ & 13.2 \\
    & \# Monosyllables & 0.08 & 0.16$^*$ & 0.39$^*$ & 0.01 & 0.22$^*$ & 18.8 \\
    & \# Polysyllables & 0.31$^*$ & -0.33$^*$ & 0.46$^*$ & 0.47$^*$ & 0.41$^*$ & 9.6\\
    & \# Difficult Words & 0.26$^*$ & -0.40$^*$ & 0.45$^*$ & 0.46$^*$ & \textbf{0.48}$^*$ & 8.6\\
    & TE Score & 0.35$^*$ & -0.18$^*$ & 0.34$^*$ & 0.34$^*$ & 0.06$^*$ & 17.2 \\
\midrule
\multirow{8}{*}{\textit{Psycholinguistics}} 
    & Automatic Readability Index & 0.07 & -0.33$^*$ & 0.36$^*$ & 0.56$^*$ & 0.40$^*$ & 11.0 \\
    & Coleman Liau Index & 0.30$^*$ & -0.32$^*$ & 0.31$^*$ & 0.54$^*$ & 0.35$^*$ & 16.8 \\
    & Dalle Chall Readability Score & 0.37$^*$ & -0.37$^*$ & 0.37$^*$ & 0.52$^*$ & 0.22$^*$ & 12.4 \\
    & Flesch Reading Grade & 0.15 & -0.36$^*$ & 0.37$^*$ & 0.58$^*$ & 0.40$^*$ & 11.6 \\
    & Flesch-Kincaid Reading Ease & -0.32$^*$ & 0.37$^*$ & -0.35$^*$ & -0.58$^*$ & -0.36$^*$ & 11.8 \\
    & Gunning Fog & 0.15$^*$ & -0.37$^*$ & 0.39$^*$ & 0.57$^*$ & 0.37$^*$ & 14.0 \\
    & Linsear Write Formula & -0.06 & -0.31$^*$ & 0.24$^*$ & 0.45$^*$ & 0.40$^*$ & 14.2 \\
    & SMOG Index & 0.14 & -0.38$^*$ & 0.37$^*$ & 0.59$^*$ & 0.37$^*$ & 12.2 \\
\midrule
\multirow{5}{*}{\textit{Model-based}}
    & \readme                   & 0.40$^*$ & \textbf{-0.45}$^*$ & \textbf{0.50}$^*$ &  0.50$^*$ & 0.44$^*$ & 6.2 \\
    & Meta Rater (readability) & -0.17 & 0.14$^*$ & 0.00 & 0.00 & 0.09$^*$ & 21.0 \\
    & Meta Rater (professionalism) & \textbf{0.49}$^*$ & -0.40$^*$ & \textbf{0.51}$^*$ & \textbf{0.67}$^*$ & 0.44$^*$ & \textbf{4.2} \\
    & LLM-as-a-judge (0-shot) & \textbf{0.57}$^*$ & \textbf{-0.50}$^*$ & \textbf{0.49}$^*$ & \textbf{0.73}$^*$ & \textbf{0.60}$^*$ & \textbf{2.4} \\
    & LLM-as-a-judge (5-shot) & \textbf{0.61}$^*$ & \textbf{-0.55}$^*$ & 0.43$^*$ & \textbf{0.71}$^*$ & \textbf{0.61}$^*$ & \textbf{3.2} \\
    & LLM-as-a-judge (continuous 0-100) & \textbf{-0.56}$^*$ & \textbf{0.59}$^*$ & \textbf{-0.53}$^*$ & \textbf{-0.68}$^*$ & \textbf{-0.52}$^*$ & \textbf{2.4} \\
\bottomrule
\end{tabular}}
\caption{\textbf{Rank correlations between readability metrics and human judgments of correctness across 5 datasets.} We report the Kendall Tau coefficient and boldface the four metrics exhibiting strongest correlations with human judgments. $^*$indicates correlation coefficients with p-value $<0.01$.}
\label{tab:results:correlation}
\postspace\postspace
\end{table*}
An ideal metric should correlate strongly with human judgments of readability. To operationalize this, and given that readability labels are ordinal, we map the discrete labels to monotonically increasing numeric values ranging from 0 to $k-1$. We apply a similar transformation to the outputs of model-based metrics to obtain numerical values and then compute the correlation between metric outputs and human annotations using the Kendall Tau-b coefficient~\citep{Kendall1938ANM}.\footnote{We use the implementation available in \texttt{scipy.stats}.}\footnote{See Appendix~\ref{appdx:automated-metrics:long-definition} for details on the categorical-to-numerical mappings used for each metric.} To assess overall performance, we report the average rank order across all datasets (\texttt{Avg. Rank}).

Table \ref{tab:results:correlation} shows that \textbf{\textit{model-based metrics systematically achieve stronger correlations with human judgments}}, surpassing surface-form and psycholinguistic metrics by up to 0.24 absolute points. Notably, all three LLM-as-a-judge metrics consistently rank in the top three (average ranks 2.4–3.2), followed closely by the fine-tuned \metaraterprofessionalism and \readme models.
Looking at the disagreements between metrics, we find LLM-as-a-judge metrics to be more sensitive to specialized terminology and sentence structure, whereas fine-tuned models like \readme are more sensitive to information density and presence of connectors and cohesive devices.
Comparing \metaraterprofessionalism with \metaraterreadability, the latter shows an average correlation rank of 21.0, falling below psycholinguistic and surface-form metrics, where the best traditional metric achieves 8.6. This may be because examples are generally clear, grammatically correct, and coherent, leading the model to systematically assign the same readability class. Conversely, because \metaraterprofessionalism reflects the depth and expertise demanded by each input, we hypothesize it better aligns with human perceptions of readability which go beyond lexical and syntactic cues (see Section \ref{sec:human-perceptions-readability}).

Together these results demonstrate the strong performance of LLM-as-a-judge metrics. However, we highlight the trade-off with inference cost, as LLM-based evaluations typically require generating text for each instance, making them slower and more resource-intensive approaches than fine-tuned models. We also note that despite achieving the strongest correlations with human judgments (up to 0.73), \textbf{\textit{model-based metrics remain far from perfect alignment}}, suggesting room for improvement.

Overall, \textbf{no single model-based metric consistently dominates}: while the continuous LLM-as-a-judge metric achieves the highest correlations on three datasets, it underperforms relative to \llmzeroshot on \eliwhyhuman and \scienceqa. The two metrics differ considerably: the continuous variant penalizes texts containing numbers and named entities (\eg ``The Barber of Seville''), whereas the discriminative one is more sensitive to scientific terminology (\eg ``hydrophobic effect'', ``endergonicity''), complex sentence structures, and equations. Despite its finer granularity, the continuous approach shows marked score saturation in \scienceqa~\citep{li2025evaluatingscoringbiasllmasajudge}, with 81.30\% of scores confined to three values.

\textbf{Surface-form metrics outperform psycholinguistic metrics on 4 (out of 5) datasets}. 
With the exception of \eliwhyhuman dataset, Table \ref{tab:results:correlation} shows that there is always a simpler surface-level metric (\eg \textsc{\# Difficult Words}, or \textsc{\#Syllables}) that is on par or outperforms popular metrics, such as the Automatic Readability Index or the Flesch Kincaid Reading Ease. Upon further analysis, we find that the stronger correlation observed for average sentence length in the \eliwhygpt can be attributed to length bias in the generations, where perceived readability is linked to the explanation's length (see Figure \ref{fig:dataset:eliwhygpt:length-bias}). 

\section{Conclusion}
\label{sec:conclusion}
\minipostspace
This work tackles the inconsistency of readability definitions (and metrics) in the literature by showing that human perceptions of readability go beyond lexical and syntactic features, also considering topic and information content.
Furthermore, we benchmark 20+ reference-less metrics--including LLM-as-a-judge and fine-tuned models--across five datasets. Our results show that model-based metrics correlate more strongly with human judgments than popular readability metrics, suggesting they capture more nuanced features. Together, these findings call for clearer definitions of readability and more rigorous validation of metrics, paving the way for assessments that better reflect how humans understand text.

\section*{Limitations}
The analysis conducted in this paper is limited to the available datasets in the English language, therefore providing limited generalization to other languages.
While we are partially motivated by the lack of high quality labeled data in other languages, a few exceptions exist namely in the French language~\citep{francois-fairon-2012-ai}.
Future work may consider expanding on this work through the creation of additional readability datasets in other languages or by expanding our analysis to other languages.

Section \ref{sec:human-perceptions-readability} concerns the investigation of the main factors shaping human readability judgments. While our findings are intuitive and generally aligned with prior discussion in the literature~\citep{august-et-al-2024,Klare1974-linsearwrite}, they are based on information extracted from a single dataset in QA, potentially leading to concerns about their generalizability. 
However, reasoning judgments are not widely available in readability datasets, making it non-trivial to extend this analysis to other datasets.
Future work could include building additional datasets, therefore, facilitating the expansion of this analysis to other domains and tasks. 

\section*{Lay Summary}

Readability assessment helps ensure that information can be understood by people with different backgrounds and abilities. A key goal is to automate this process and reduce the need for human evaluation.

Many datasets and methods have been developed for automatic readability assessment, but they often rely on different definitions of what makes text readable. Even today, most approaches still use basic measures, like the number of words, syllables, or sentences, to estimate readability.

In this work, we show that people's perceptions of readability depend on more than simple text features—they are strongly influenced by the content and topic of the text. We compare traditional readability measures with more advanced model-based metrics across five datasets and find that conventional measures often fail to capture what humans consider readable. Our results emphasize the need for clearer, standardized definitions of readability and for moving beyond simple, surface-level metrics.

\section*{Acknowledgments}
We thank the anonymous reviewers, the members of the Capital One research team for their helpful feedback.

\bibliography{acl}
\appendix

\section{Additional Details: Datasets}
\label{appdx:dataset:statistics}

In this section, we provide additional details about the datasets. 
Table \ref{tab:datasets-statistics} summarizes the general statistics about the five datasets considered in this study, including the readability label type, the size of the dataset, but also the average example length in terms of word count and sentence count. 

\subsection{\scienceqa~\citep{lu2022learn-scienceqa}}

\scienceqa is collected from elementary and high school science curricula sourced from IXL learning\footnote{https://www.ixl.com} and with topics ranging from natural, social, and natural sciences. 
To ensure coverage across grades 1–12, we sample from the full dataset. 
We draw 200 examples per grade, except for 1st grade where only 95 are available, yielding 2295 examples overall. 
Although primarily a multiple-choice QA dataset, it also includes a \textit{lecture} covering the knowledge needed to answer each question and a solution outlining how to use it to derive the answer. 
For every question, we compute the readability by concatenating the two fields as demonstrated in Figure~\ref{fig:dataset-details:scienceqa-example-format}. 
For some qualitative examples, see Table~\ref{tab:qualitative-examples:dataset:scienceqa}.
To compute the correlation with human judgments, we use grades 1-12 as the readability judgments (12-way classification), where a higher grade implies added difficulty in comprehending a text.
\begin{figure}[tb]
\centering
\small
\begin{tcolorbox}[fonttitle=\fontfamily{pbk}\selectfont\bfseries,
                  fontupper=\fontsize{8}{9}\fontfamily{ppl}\selectfont\itshape,
                  fontlower=\fontfamily{put}\selectfont\scshape,
                  title=\scienceqa readability example,
                  width=\linewidth,
                  arc=1mm, auto outer arc]
\begin{Verbatim}[breaklines=true, breaksymbol={}]
Lecture: {{lecture}}
Explanation: {{explanation}}
\end{Verbatim}
\end{tcolorbox}
\postspace
\minipostspace
\caption{Formatting of each \scienceqa example. Whenever examples miss the corresponding \texttt{\{\{lecture\}\}} or \texttt{\{\{explanation\}\}} fields, we we omit them from the template above.} 
\label{fig:dataset-details:scienceqa-example-format}
\end{figure}

\begin{table*}[tb]
\small
\centering
\caption{Randomly selected ScienceQA examples across 6 different readability classes (\textbf{grade}).}
\label{tab:qualitative-examples:dataset:scienceqa}
\begin{tabular}{c p{0.15\linewidth}  p{0.70\linewidth}}
\toprule
\textbf{Grade}  & \textbf{Subject \newline(Category)} & \textbf{Formatted Example} \\
\midrule
\rowcolor{gray!10}  1   & language science \newline (comprehension strategies) & Explanation: A book is made of paper.\n A book tells a story.\n A teacher may read a book out loud.\\
\addlinespace 
\rowcolor{gray!0} 3 & natural science \newline (weather and climate)  & Lecture: The atmosphere is the layer of air that surrounds Earth. Both weather and climate tell you about the atmosphere.\n Weather is what the atmosphere is like at a certain place and time. Weather can change quickly. For example, the temperature outside your house might get higher throughout the day.\n Climate is the pattern of weather in a certain place. For example, summer temperatures in New York are usually higher than winter temperatures.\n \n Explanation: Read the text carefully.\n Where Sarah lives, winter is the rainiest season of the year.\n This passage tells you about the usual precipitation where Sarah lives. It does not describe what the weather is like on a particular day. So, this passage describes the climate. \\ \addlinespace
\rowcolor{gray!10} 5 & natural science \newline (traits and heredity) & Lecture: Organisms, including people, have both inherited and acquired traits. Inherited and acquired traits are gained in different ways.\n Inherited traits are passed down through families. Children gain these traits from their parents. Inherited traits do not need to be learned.\n Acquired traits are gained during a person's life. Some acquired traits, such as riding a bicycle, are gained by learning. Other acquired traits, such as scars, are caused by the environment. Children do not inherit their parents' acquired traits.\n \n Explanation: People are not born knowing how to cook. Instead, many people learn how to cook. So, cooking is an acquired trait. \\
\addlinespace
\rowcolor{gray!0} 7 & natural science \newline (designing experiments) & Lecture: Experiments can be designed to answer specific questions. When designing an experiment, you must identify the supplies that are necessary to answer your question. In order to do this, you need to figure out what will be tested and what will be measured during the experiment.\n Imagine that you are wondering if plants grow to different heights when planted in different types of soil. How might you decide what supplies are necessary to conduct this experiment?\n First, you need to identify the part of the experiment that will be tested, which is the independent variable. This is usually the part of the experiment that is different or changed. In this case, you would like to know how plants grow in different types of soil. So, you must have different types of soil available.\n Next, you need to identify the part of the experiment that will be measured or observed, which is the dependent variable. In this experiment, you would like to know if some plants grow taller than others. So, you must be able to compare the plants' heights. To do this, you can observe which plants are taller by looking at them, or you can measure their exact heights with a meterstick.\n So, if you have different types of soil and can observe or measure the heights of your plants, then you have the supplies you need to investigate your question with an experiment! \\ \addlinespace
\rowcolor{gray!10} 9 & language science \newline (literary devices) & Lecture: Figures of speech are words or phrases that use language in a nonliteral or unusual way. They can make writing more expressive.\n A euphemism is a polite or indirect expression that is used to de-emphasize an unpleasant topic.\n The head of Human Resources would never refer to firing people, only to laying them off.\n Hyperbole is an obvious exaggeration that is not meant to be taken literally.\n I ate so much that I think I might explode!\n An oxymoron is a joining of two seemingly contradictory terms.\n Some reviewers are calling this book a new classic.\n A paradox is a statement that might at first appear to be contradictory, but that may in fact contain some truth.\n Always expect the unexpected.\n \n Explanation: The text uses an oxymoron, a joining of two seemingly contradictory terms.\n Open secret is a contradiction, because open describes something that is freely or publicly known, and a secret is hidden. \\      
\bottomrule
\end{tabular}
\end{table*}

\begin{table*}[tb]
\ContinuedFloat
\small
\centering
\caption{Randomly selected ScienceQA examples across 6 different readability classes (\textbf{grade}). (continued)}
\begin{tabular}{c p{0.15\linewidth} p{0.70\linewidth}}
\toprule
\textbf{Grade} & \textbf{Subject \newline(Category)} & \textbf{Formatted Example} \\
\midrule
\rowcolor{gray!0} 11 & language science (word usage and nuance) & Lecture: Words change in meaning when speakers begin using them in new ways. For example, the word peruse once only meant to examine in detail, but it's now also commonly used to mean to look through in a casual manner.\n When a word changes in meaning, its correct usage is often debated. Although a newer sense of the word may be more commonly used, many people consider a word's traditional definition to be the correct usage. Being able to distinguish the different uses of a word can help you use it appropriately for different audiences.\n Britney perused her notes, carefully preparing for her exam.\n The traditional usage above is considered more standard.\n David perused the magazine, absentmindedly flipping through the pages.\n The nontraditional usage above is now commonly used, but traditional style guides generally advise against it.\n \n 
Explanation: The first text uses travesty in its traditional sense: a ridiculous imitation; a parody.\n Doug's ill-researched essay about the Space Race received a poor grade because it presented such a travesty of the actual historical events.\n The second text uses travesty in its nontraditional sense: a disappointment or a tragedy.\n Doug realized that his essay about the Space Race was a bit inaccurate, but he still thought it a travesty that such an entertaining essay should receive a poor grade.\n Most style guides recommend to use the traditional sense of the word travesty because it is considered more standard. \\     
\bottomrule
\end{tabular}
\end{table*}

\subsection{\cleardataset~\citep{crossley-et-al-2023-CLEAR-dataset}}

\cleardataset consists of 4.7k text excerpts sampled from online digital libraries.
Each example is curated to ensure the text is self-contained and composed of full sentences. 
Unlike the other datasets, the readability score in \cleardataset is continuous and represents the easiness of comprehension of a given text (\textit{BT\_easiness}). 
We refer to the original paper for additional details regarding the dataset.
Table \ref{tab:qualitative-examples:dataset:clear} illustrates a few examples from this dataset and corresponding readability score.
To balance efficiency with generalization, we randomly sample 1k examples without replacement from the original dataset and use them for our correlation analysis. 
Table \ref{tab:statistics:dataset:clear} 
\begin{table*}[tb]
\centering
\caption{Randomly selected examples from the \cleardataset dataset. In contrast to other datasets, each example is associated with a continuous readability score obtained by fitting a Bradley–Terry model on pairwise human judgments of reading ease.}
\label{tab:qualitative-examples:dataset:clear}
\small
\begin{tabular}{p{0.10\linewidth} p{0.06\linewidth} p{0.77\linewidth}}
\toprule
\textbf{Readability Score} & \textbf{Category} & \textbf{Text} \\
\midrule
\rowcolor{gray!10} -2.91 & Info & It must not be supposed that, in setting forth the memories of this half-hour between the moment my uncle left my room till we met again at dinner, I am losing sight of "Almayer's Folly." Having confessed that my first novel was begun in idleness--a holiday task--I think I have also given the impression that it was a much-delayed book. It was never dismissed from my mind, even when the hope of ever finishing it was very faint. Many things came in its way: daily duties, new impressions, old memories. It was not the outcome of a need--the famous need of self-expression which artists find in their search for motives. The necessity which impelled me was a hidden, obscure necessity, a completely masked and unaccountable phenomenon. Or perhaps some idle and frivolous magician (there must be magicians in London) had cast a spell over me through his parlour window as I explored the maze of streets east and west in solitary leisurely walks without chart and compass. Till I began to write that novel I had written nothing but letters, and not very many of these. \\
\rowcolor{gray!00} -1.44 & Info & In the second place, the Emperor is an exceedingly intelligent and highly cultivated man. His mental processes are swift, but they go also very deep. He is a searching inquirer, and questions and listens more than he talks. His fund of knowledge is immense and sometimes astonishing. He manifests interest in everything, even to the smallest detail, which can have any bearing upon human improvement. I remember a half hour's conversation with him once over a cupping glass, which he had gotten from an excavation in the Roman ruin called the Saalburg, near Homburg. He always appeared to me most deeply concerned with the arts of peace. I have never heard him speak much of war, and then always with abhorrence, nor much of military matters, but improved agriculture, invention, and manufacture, and especially commerce and education in all their ramifications, were the chief subjects of his thought and conversation. I have had the privilege of association with many highly intelligent and profoundly learned men, but I have never acquired as much knowledge, in the same time, from any man whom I have ever met, as from the German Emperor. \\
\rowcolor{gray!10}  -1.21 & Literary & Moreover Grandmother Grant always dressed in one fashion; she had a calico dress for morning and a black silk for the afternoon, made with an old-fashioned surplice waist, with a thick plaited ruff about her throat; she sometimes tied a large white apron on, but only when she went into the kitchen; and she wore a pocket as big as three of yours, Matilda, tied on underneath and reached through a slit in her gown. Therein she kept her keys, her smelling-bottle, her pocket-book, her handkerchief and her spectacles, a bit of flagroot and some liquorice stick. I mean when I say this, that all these things belonged in her pocket, and she meant to keep them there; but it was one peculiarity of the dear old lady, that she always lost her necessary conveniences, and lost them every day.\n"Maria!" she would call out to her daughter in the next room, "have you seen my spectacles?"\n"No, mother; when did you have them?"\n"Five minutes ago, darning Harry\'s stockings; but never mind, there\'s another pair in the basket." \\
\rowcolor{gray!00} -0.37 & Literary & The others were watching him closely. They guessed something of the nature of what must be passing through Ned's mind, for both Jack and Teddy followed his gaze up the uneven shore. Jimmy had the glasses again, and was busily engaged in scrutinizing the blur on the distant horizon, which all of them had agreed must be smoke hovering close to the water. Perhaps he half-believed the fanciful suggestion made by Teddy, with reference to Captain Kidd, and was wildly hoping to discover some positive sign that would stamp this fairy story with truth. All the previous adventures that had befallen himself and chums would sink into utter insignificance, could they go back home and show evidences of having made such a romantic discovery up there in the Hudson Bay country.\n"See the feather they say he always wore in his hat, Jimmy?" asked Frank. \\
\rowcolor{gray!10} 0.06 & Literary & The other day, as I was walking through a side street in one of our large cities, I heard these words ringing out from a room so crowded with people that I could but just see the auctioneer's face and uplifted hammer above the heads of the crowd.\n"Going! Going! Going! Gone!" and down came the hammer with a sharp rap.\n I do not know how or why it was, but the words struck me with a new force and significance. I had heard them hundreds of times before, with only a sense of amusement. This time they sounded solemn.\n"Going! Going! Gone!"\n"That is the way it is with life," I said to myself - "with time." This world is a sort of auction room; we do not know that we are buyers: we are, in fact, more like beggars; we have brought no money to exchange for precious minutes, hours, days, or years; they are given to us. There is no calling out of terms, no noisy auctioneer, no hammer; but nevertheless, the time is "going! going! gone!" \\
\bottomrule
\end{tabular}
\end{table*}
\begin{table*}[tb]
\centering
\caption{Randomly selected examples from the \cleardataset dataset. In contrast to other datasets, each example is associated with a continuous readability score obtained by fitting a Bradley–Terry model on pairwise human judgments of reading ease. (continued)}
\label{tab:qualitative-examples:dataset:clear}
\small
\begin{tabular}{p{0.10\linewidth} p{0.06\linewidth} p{0.77\linewidth}}
\toprule
\textbf{Readability Score} & \textbf{Category} & \textbf{Text} \\
\midrule
\rowcolor{gray!00} 0.19 & Info & There are various kinds of pitcher-plants. Some are shorter and broader than others; but they are all green like true leaves, and hold water as securely as a jug or glass. They grow in Borneo and Sumatra, hot islands in the East. The one shown in the drawing grows in Ceylon.\n Some grow in America; but they are altogether different from those in Borneo and Ceylon. One beautiful little pitcher-plant grows in Australia: but this is also very different from all the rest; for the pitchers, instead of being at the end of the leaves, are clustered round the bottom of the plant, close to the ground.\n All these pitcher-plants, though very beautiful to look at, are very cruel enemies to insects: for the pitchers nearly always have water in them; and flies and small insects are constantly falling into them, and getting drowned. \\
\bottomrule
\end{tabular}
\end{table*}

\begin{table}[tb]
\centering
\caption{Comparison of readability scores between the original \cleardataset (Original) and the 1k subsample used to conduct the correlation analysis (Subsample).}
\label{tab:statistics:dataset:clear}
\small
\begin{tabular}{c c c}
\toprule
\textbf{Statistic}  & \textbf{Original} & \textbf{Subsample} \\
\midrule
Count   &  4724 & 1000 \\
Mean    & -0.96 & -0.97 \\
Std     & 1.03  &  1.06 \\
Min     & -3.68 & -3.68 \\ 
25\%    & -1.70 & -1.74\\
50\%    & -0.91 & -0.89 \\
75\%    & -0.20 & -0.20\\
Max     & 1.71  &  1.71 \\
\bottomrule
\end{tabular}
\end{table}

\subsection{\augustetal~\citep{august-et-al-2023-paper-plain}}

\augustetal dataset is a parallel corpus for readability, comprising 3 human-edited variants of the same summary for each example. 
Table~\ref{tab:qualitative-examples:dataset:august-et-al} shows three human-curated versions of the question ``What did the paper find?'' at different complexity levels.
The correlation analysis considers we all examples and map the ordinal classes—\lowcomplexity $\prec$ \mediumcomplexity $\prec$ \highcomplexity—onto a 0–2 scale.

\begin{table*}[tb]
\centering
\caption{Randomly selected examples from the \augustetal dataset, spanning all three readability classes.}
\label{tab:qualitative-examples:dataset:august-et-al}
\small
\begin{tabular}{l p{0.75\linewidth}}
\toprule
\textbf{Complexity Level} & \textbf{Text} \\
\midrule
\rowcolor{gray!10} \lowcomplexity & The researchers found that women who lived in countries that received less US foreign aid during the policy used less contraceptives and had both more pregnancies and more abortions during the years that the policy was in place. They also noted that the effects of the policy reversed once it had been rescinded, further strengthening the researchers’ hypothesis that the Mexico City Policy has an effect on a nation’s observed patterns of reproductive behavior.\\ \addlinespace 
\rowcolor{gray!00} \mediumcomplexity & The researchers found that abortions and pregnancies increased when the Mexico City Policy was in effect, which they correlate to a decreased availability in contraception during those years. They also found that the effects varied by exposure to the policy, as women in high exposure countries were more likely to experience abortion when the policy was enacted and less likely when it wasn’t in effect. The alternating patterns of reproductive behavior depending on whether the policy was enacted also strengthens the researchers’ hypothesis that it has a not insubstantial effect on abortion rates in sub-Saharan Africa. \\ \addlinespace 
\rowcolor{gray!10} \highcomplexity & When US support for international family planning organizations was conditioned on the policy, coverage of modern contraception fell and the proportion of women reporting pregnancy and abortions increased, in relative terms, among women in countries more reliant on US funding. Although the degree to which abortions increase when contraceptive supply is curtailed is poorly characterized, one analysis estimated that, depending on the total fertility of the population, a 10\% decline in contraceptive use would lead to a 20-90\% increase in abortions. The researchers posit that the observed changes in abortion could be due to changing availability of modern contraception, and that a change in the use of modern\r\n contraception would be expected to result in a change in pregnancy rates. Women in high-exposure countries experienced a relative increase in abortion (and decrease in modern contraceptive use) when the policy was enacted and a relative decrease in abortion (and increase in modern contraceptive use) when the policy was rescinded.\\
\addlinespace
\hdashline \hdashline
\addlinespace
\rowcolor{gray!00} \lowcomplexity & Study looks at pushup capacity and heart health, finding that those who could do the most (over 40) push ups had the lowest risk of heart disease. \\ \addlinespace
\rowcolor{gray!10}  \mediumcomplexity &  Study examines the relationship between a person’s push up ability and their physical health, finding that push ups are a good indicator of a person’s cardiovascular fitness. \\ \addlinespace
\rowcolor{gray!00} \highcomplexity & Association Between Pushup Exercise Capacity and Future Cardiovascular Events Among Active Adult Men \\
\bottomrule
\end{tabular}
\end{table*}

\subsection{\eliwhygpt~\citep{joshi2025eliwhyevaluatingpedagogicalutility}}
\begin{figure*}[tb]
  \includegraphics[width=0.49\linewidth]{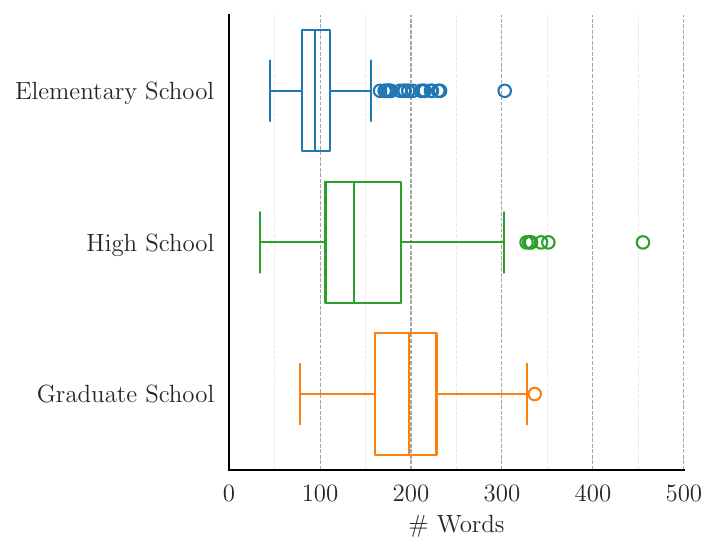} \hfill
  \includegraphics[width=0.49\linewidth]{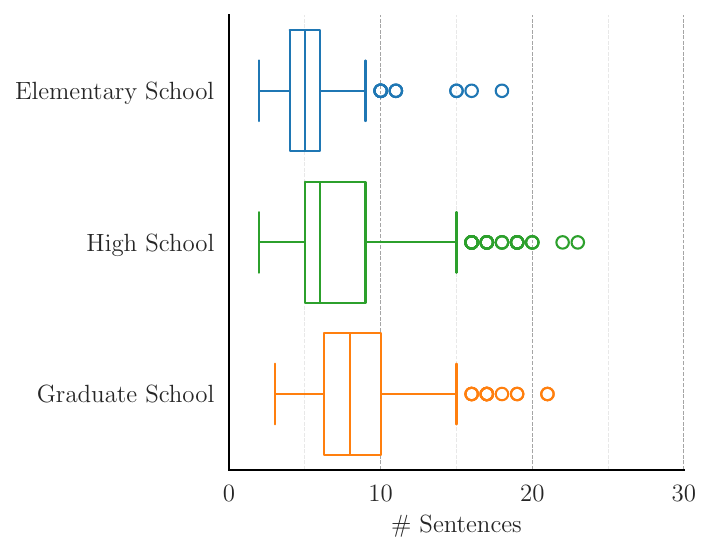}
  \caption {Distribution of number of words (\textsc{\# Words}) and sentences (\textsc{\# Sentences}) per readability label in the \eliwhygpt dataset.}
  \label{fig:dataset:eliwhygpt:length-bias}
\end{figure*}

Our analyses reveal the presence of length bias, where there seems to exist a correlation between the length of GPT4-generated explanations and human perceived readability (see Figure \ref{fig:dataset:eliwhygpt:length-bias}). 
In fact, we observe a propensity for responses deemed higher readability to be longer, which can be explained by the added detail and specificity often emphasized by human experts.
Future work could explore ways of mitigating this bias by enforcing strict generation lengths or, if a reference document with relevant information is available by controlling the information content within each generation~\citep{august-et-al-2024}.

\subsection{\eliwhyhuman~\citep{joshi2025eliwhyevaluatingpedagogicalutility}}

Table \ref{tab:qualitative-examples:dataset:eliwhy-human} illustrates a few randomly selected examples for the \eliwhyhuman datasets. These explanations were manually curated by two authors of the paper.

\begin{table*}[tb]
\centering
\caption{Examples of different explanations for the \eliwhyhuman for the questions ``Why do we enjoy horror movies or stories?'' and ``'Why does DNA have a double helix structure?''. Each set of three examples refers to the same question. }
\label{tab:qualitative-examples:dataset:eliwhy-human}
\small
\begin{tabular}{p{0.10\linewidth} p{0.10\linewidth} p{0.70\linewidth}}
\toprule
\textbf{Readability}  & \textbf{Topic} & \textbf{Formatted Explanation} \\
\midrule
\rowcolor{gray!10}  \esshort & Psychology & All the same reasons people like sad songs, Halloween, war documentaries, apocalyptic fiction, etc. etc. It’s like any other film genre. Horror movies can be artistic; the performances can be entertaining; the movies can be well-constructed or conceived; they can be relatable or provide personal insight. \\
\rowcolor{gray!00}  \hsshort & Psychology & According to these researchers, stimulation is one of the driving forces behind the consumption of horror. Exposure to terrifying acts like stories of demonic possession or alien infestation can be stimulating both mentally and physically. These experiences can give rise to both negative feelings, such as fear or anxiety, and positive feelings, such as excitement or joy. And we tend to feel the most positive emotions when something makes us feel the most negative ones. \\
\rowcolor{gray!10}   \phdshort & Psychology & “The horror film occupies in popular culture roughly comparable to that of horror literature. That is to say, it is generally ignored, sometimes acknowledged with bemused tolerance, and viewed with alarm when it irritates authority - rather like a child too spirited to follow the rules that rendition has deemed acceptable” (p. ix), a view that is echoed elsewhere. For example, Tudor (1997) noted that “a taste of horror is a taste for something seemingly abnormal and is therefore deemed to require special attention” (p. 446). Part of the reason for the disdain, apart from the broad and base nature of the content, may be the relative cheapness of horror film: these are often much less expensive to create than are other genre films such as westerns, comedies, or science fiction. \\
\hdashline \hdashline
\rowcolor{gray!00}  \esshort & Biology & DNA is made up of small components called nucleotides. A nucleotide is made up of 3 parts: a phosphate group, a sugar, and a base. The base can be 1 of 4 varieties: A, T, C, or G. Simply put, because of their structures, A and T bond nicely with one another and C and G bond nicely as well. As a result, a single strand of DNA will bond to a strand with another strand with a "complimentary" sequence of bases. In other words, there will be two strands with opposite, for lack of a better word, sequences of bases. The production of strands that complement one another is a result of how DNA copies itself (known as DNA replication). The second strand's base sequence is ordered based on the first strand\'s sequence.\n\n So that answers the question of why DNA is a double helix instead of a single helix. But why is it a helix at all? Why not a shaped like a ladder? In a cell, most of the material present is water. Water is shaped in such a way where positive and negative charges aren't spread evenly throughout the water\'s molecules. This is known as being a polar molecule. Polar molecules "like" being around other polar molecules. Non-polar molecules don't "like" being around polar molecules like water. In DNA, the bases are non-polar, but the phosphate groups are polar. As a result, the preferred shape puts the phosphates in contact with the water and the bases are covered by being on the inside. The twisting shape of DNA reduces the extent to which the bases are exposed to the water in the outside environment.\n\n TL;DR: DNA is made of 3 components: bases, sugars, and phosphates. The sugars bond with one another explaining the double-configuration of DNA. To be stable, the bases need to stay in the inside of the molecule and phosphates need to make up the molecule\'s exterior, and a helical shape allows this to happen.\\
\rowcolor{gray!10}  \hsshort & Biology & The three-dimensional structure of DNA—the double helix—arises from the chemical and structural features of its two polynucleotide chains. Because these two chains are held together by hydrogen bonding between the bases on the different strands, all the bases are on the inside of the double helix, and the sugar-phosphate backbones are on the outside. To maximize the efficiency of base-pair packing, the two sugar-phosphate backbones wind around each other to form a double helix, with one complete turn every ten base pairs. \\
\rowcolor{gray!00} \phdshort & Biology & The stability of double-stranded DNA (dsDNA) at physiological temperature is due to the hydrogen bonding between complementary bases and the stacking between neighboring bases. However, these base-stacking interactions are of the order of magnitude of a few k\_B T thermal energy and the thermal fluctuations can lead (even at physiological temperature) to local and transient unzipping of the double helix. \\
\bottomrule
\end{tabular}
\end{table*}

\section{Automated Metrics}
\label{appdx:automated-metrics:long-definition}
This section discusses the implementation details of the metrics evaluated in the main paper. All experiments are implemented in Python: 
\texttt{textstat}\footnote{\url{https://pypi.org/project/textstat/}} is used to compute surface-form and psycholinguistic metrics; 
\texttt{transformers} is used to implement the model-based metrics, including fine-tuned and LLM-as-a-judge approaches.

\subsection{Psycholinguistics Metrics}
\label{appdx:automated-metrics:ssec:psycholinguistic-metrics}
The metrics listed below are commonly referred to as \textit{readability tests} and commonly used to gauge the difficulty that human readers may have in understanding a given text. 

\textbf{Automatic Readability Index (ARI)}~\citep{senter1967automated} estimates the US grade level needed to comprehend a text. 
To do so, it uses the ratio of characters-to-words and words-to-sentences. 
Intuitively, these ratios capture the idea that longer words and longer sentences are more difficult to grasp. 
The character counts include both numbers and letters. 
A score of 1 and 14 would match that of a Kindergarten and a College student, respectively.
\begin{equation*}
  \label{eq:metric:automatic-readability-index}
  \Scale[0.98]{\left\lceil 4.71\left(\frac{\texttt{\#chars}}{\texttt{\#words}}\right)  + 0.5\left(\frac{\texttt{\#words}}{\texttt{\#sentences}}\right) - 21.43 \right\rceil}
\end{equation*}

\textbf{Coleman Liau Index (CLI)}~\citep{coleman1975computer} similarly to ARI, it also yields an estimate of the minimum US grade level necessary to understand a piece of text. It is defined in terms of the average counts of letters and sentences per 100 words in a text sample. 
\begin{equation*}
  \label{eq:metric:coleman-liu-index}
    \Scale[0.98]{0.0588 \cdot \texttt{\#letters} - 0.296 \cdot \texttt{\#sentences} - 15.8}
\end{equation*}

\textbf{Dale Chall Readability}~\citep{dalle-chall-readability} leverages the fraction of difficult words in the document, as well as the average word-to-sentence count ratio to gauge the difficulty of a given text. 
By design, the metric relies on a pre-defined subset of 3k words that is empirically expected to be familiar to the majority of 4th graders. 
The formula is designed such that scores $\leq 4.9$ match grade 4 and below, and scores $\geq 10$ match grades 16 and above. 
Below we write the new Dale-Chall Formula:
\begin{equation*}
  \label{eq:metric:dale-chall-readability}
  \Scale[0.98]{\left\lfloor 64 - 0.95\left(\frac{\texttt{\#difficult\_words}}{\texttt{\#words}}\right)  - 0.69\left(\frac{\texttt{\#words}}{\texttt{\#sentences}}\right) \right\rfloor}
\end{equation*}

\textbf{Flesch-Kincaid Reading Ease (FKRE)} and \textbf{Flesch-Kincaid Grade Level (FKGL)}~\citep{Flesch1948ANR} rely on the same core properties of language, such as average word length and average sentence length, differing only in the coefficients. 
The formulas were defined by the US Navy to gauge the readability of the technical material and later adopted by a few US states to impose readability requirements on various legal documents (\eg insurance policies)~\citep{readability-metric-useful}.
The FKRE is defined in as follows:
\begin{equation*}
  \label{eq:metric:fkre}
  \Scale[0.98]{206.835 - 1.015 \left( \frac{\texttt{\#words}}{\texttt{\#sentences}} \right) - 84.6 \left( \frac{\texttt{\#syllables}}{\texttt{\#words}} \right)}
\end{equation*}
whereas the FKGL is defined as:
\begin{equation*}
  \label{eq:metric:fkgl}
  \Scale[0.98]{0.39 \left( \frac{\texttt{\#words}}{\texttt{\#sentences}} \right) + 11.8 \left( \frac{\texttt{\#syllables}}{\texttt{\#words}} \right) - 15.59}
\end{equation*}

\textbf{Gunning Fog Index (GFI)}~\citep{gunning1952technique} 
provides an estimate of the number of formal education required to understand the text on a first reading. 
It works by first computing the average sentence length, \ie word-to-sentence ratio of a passage and then computing the ratio of complex words in the passage. 
In this formula, complex words are defined as words with 3+ syllables that are not proper nouns, familiar words, or compound words. 
Conventionally, scores range between 6 and 17 which indicate that 6th grade and College Graduate are necessary to be able to understand a piece of text, respectively.
\begin{equation*}
  \label{eq:metric:gunning-fog}
  \Scale[0.98]{0.4\left[\left(\frac{\texttt{\#words}}{\texttt{\#sentences}}\right) + 100 \left(\frac{\texttt{\#complex\_words}}{\texttt{\#words}}\right) \right]}
\end{equation*}

\textbf{Linsear Write Formula (LWF)}~\citep{Klare1974-linsearwrite} counts the number of easy and hard words in a 100-word sample. 
To distinguish easy from hard words, it utilizes the number of syllables in each word: polysyllable words are considered hard words, whilst words with less than 3 syllables are considered easy.
It was originally designed to gauge the readability of the technical manuals used in the US Air Force.
\begin{equation*}
  \label{eq:metric:linsear-write-formula1}
  \Scale[0.98]{r = \frac{ 3 \cdot \texttt{\#hard\_words} + 1 \cdot \texttt{\#easy\_words} }{\texttt{\#words}}}
\end{equation*}
where the final linsear write score is given by
\begin{equation*}
  \label{eq:metric:linsear-write-formula2}
  \text{LWF} =
  \begin{cases}
    \dfrac{r}{2},      & \text{if } r > 20 \\[4pt]
    \dfrac{r}{2} - 1,  & \text{otherwise.}
  \end{cases}
\end{equation*}

\textbf{SMOG grade}~\citep{smog-index} 
was proposed as a more accurate and easier to compute alternative to Gunning Fog Index. 
It is defined in terms of polysyllable counts (words with 3+ syllables) across three 10-sentence long texts.
\begin{equation*}
  \label{eq:metric:smog-index}
  \Scale[0.98]{1.043\sqrt{\texttt{\#polysyllables} \cdot \frac{30}{\texttt{\#sentences}}} + 3.1291}
\end{equation*}

\subsection{Model-based Metrics}
\label{appdx:automated-metrics:ssec:model-based-metrics}

\textbf{\metaraterprofessionalism} and \textbf{\metaraterreadability}~\citep{zhuang-etal-2025-meta} are two fine-tuned based metrics, both operationalized using a ModernBERT-base model. 
The models are designed to evaluate the \textit{degree of required expertise} and \textit{ease of understanding} in a 0-5 point scale, respectively. 
To obtain the metric score associated with a given text, each text is fed through the model and the class with maximum probability is selected (\ie greedy prediction). 
This score is then used to compute the correlation with human judgments. 

\textbf{\readme}~\citep{naous-etal-2024-readme} is a model-based metric that grounds readability assessment in the capabilities of second-language learners. Specifically, we use \model{tareknaous/readabert-en}, a BERT-based model fine-tuned on the English portion of the README++ corpus---a sentence-level readability dataset spanning multiple domains (\eg finance, economics, poetry, agriculture). Readability scores are provided on a six-point scale aligned with the Common European Framework of Reference for Languages (CEFR), where higher values indicate greater language proficiency.

Since \readme was originally trained on single sentences, we hypothesize that it may not generalize well to multi-sentence inputs, such as those in \scienceqa or \eliwhygpt. To address this limitation, we adopt a bottom-up approach: for each document, we first compute the \readme score for each sentence, then average them to obtain a document-level score (\readmeavg). We also evaluate another variant, \readmemax, which reflects the hypothesis that advanced readers can understand simpler texts, but not vice versa. Table \ref{tab:results:correlation-readme-variants} summarizes the results. While both \readme and \readmemax exhibit the same average rank (1.8), we observe that \readme exhibits stronger correlations with human judgments in 3 (out of 5) evaluated datasets. Notably, \readmeavg exhibits a an average rank of 2.4, suggesting that this variant systematically under performs the other two variants in terms of correlating with human judgments. For brevity, and because of its superior performance, we restrict the analysis in the main paper to the original method---\readme.
\begin{table*}[htb]
\resizebox{\linewidth}{!}{
\begin{tabular}{l ccccc r}
\toprule
\textbf{Metric}
    & \thead{\textbf{\augustetal}\\\citep{august-et-al-2024}}  
    & \thead{\textbf{\cleardataset}\\\citep{crossley-et-al-2023-CLEAR-dataset}} 
    & \thead{\textbf{\eliwhygpt}\\\citep{joshi2025eliwhyevaluatingpedagogicalutility}} 
    & \thead{\textbf{\eliwhyhuman}\\\citep{joshi2025eliwhyevaluatingpedagogicalutility}} 
    & \thead{\textbf{\scienceqa}\\\citep{lu2022learn-scienceqa}} 
    & \thead{\textbf{Avg.} \\ \textbf{Rank}}\\
\midrule
    \readme          & \textbf{0.40} & -0.45 & \textbf{0.50} &  0.50 & \textbf{0.44} & 1.8\\
    \readmeavg       & 0.23 & -0.49 & 0.26 &  \textbf{0.68} & 0.38 & 2.4\\
    \readmemax       & 0.35 & \textbf{-0.51} & 0.43 &  0.57 & 0.42  & 1.8\\
\bottomrule
\end{tabular}}
\caption{Rank correlations between variants of the \readme metric and human judgments of correctness across 5 datasets. We boldface the variant exhibiting strongest correlation with human judgments. We report the Kendall Tau coefficient. All correlation coefficients are statistically significant with p-value $<0.01$.}
\label{tab:results:correlation-readme-variants}
\end{table*}

\textbf{\llmzeroshot} and \textbf{\llmfiveshot} are prompt-based strategies to extract the readability level from any text. Specifically, we use \model{Llama-3.3-70B-Instruct}~\footnote{\url{https://huggingface.co/meta-llama/Llama-3.3-70B-Instruct}}, a popular open-source instruction-following model.
To ensure that models' predictions strongly align with human readability judgments, we re-use instructions previously provided to humans~\citep{joshi2025eliwhyevaluatingpedagogicalutility}. 
The prompt is discriminative in nature, being designed to extract 3-way readability labels --\esshort, \hsshort, \phdshort. 
Originally, the prompt includes 5 examples of readability judgments spanning the three classes, which we refer to as \llmfiveshot. 
Although these examples improve alignment with human judgments and help constrain output structure, they add runtime overhead. 
We therefore evaluate a 0-shot version that replaces examples with explicit format instructions.
Since our goal is to compute correlations with human judgments, which can be expressed as either categorical or continuous, we map textual labels to numbers. 
Treating the labels as ordinal (\esshort $\prec$ \hsshort $\prec$ \phdshort), we assign them a 0–2 scale for correlation analysis.

\textbf{\llmshotcontinuous}, first proposed in \citet{trott-riviere-2024-measuring}, elicits continuous 0-100 readability scores from \model{GPT-4-Turbo} and \model{GPT-4o-mini}~\citep{openai2024gpt4technicalreport}, with higher values denoting easier texts to understand. 
We follow the same prompt as in the original paper (Figure~\ref{fig:prompt:llm-as-a-judge:continuous-0-100}), but replace the model with \model{Llama-3.3-70B-Instruct} to ensure comparability among LLM-as-a-judge metrics.\footnote{\model{Llama-3.3-70B-Instruct} consistently generates a number between 1–100.} 

In the main paper, we ensure the reproducibility of LLM-as-a-judge evaluations by reporting correlations obtained from greedy generations (\texttt{temperature=0}).\footnote{Continuous LLM-as-a-judge approaches (\llmshotcontinuous) are configured to generate at most 3 tokens, whereas the discriminative approaches (\llmzeroshot and \llmfiveshot) are configured to generate at most 20 tokens. We then extract the corresponding readability label through the use of regular expressions.} This decoding strategy is not only deterministic but also commonly adopted in prior work~\citep{trott-riviere-2024-measuring,gu2025surveyllmasajudge}, being representative of the most likely (or modal) behavior of the LLM.

\begin{figure}[tb]
\centering
\small
\begin{tcolorbox}[fonttitle=\fontfamily{pbk}\selectfont\bfseries,
                  fontupper=\fontsize{8}{9}\fontfamily{ppl}\selectfont\itshape,
                  fontlower=\fontfamily{put}\selectfont\scshape,
                  title=\llmshotcontinuous,
                  width=\linewidth,
                  arc=1mm, auto outer arc]
\begin{Verbatim}[breaklines=true, breaksymbol={}]
{
    "content": "You are an experienced teacher skilled at identifying the readability of different texts.",
    "role": "system"
}, {
    "content": "Read the text below. Then indicate the readability of the text, on a scale from 1 (extremely challenging to understand) to 100 (very easy to read and understand). In your assessment, consider factors such as sentence structure, vocabulary complexity, and overall clarity.\n<Text>{{text}}</Text>\nOn a scale from 1 (extremely challenging to understand) to 100 (very easy to read and understand), how readable is this text? Please answer with a single number.",
    "role": "user"
}
\end{Verbatim}
\end{tcolorbox}
\postspace
\minipostspace
\caption{Prompt used to extract a 0-100 continuous score associated with the ease of readability of a given text. The placeholder \texttt{\{\{text\}\}} is either the explanation to a question or the text excerpts depending on the dataset being evaluated.} 
\label{fig:prompt:llm-as-a-judge:continuous-0-100}
\end{figure}

\section{Human Perceptions of Readability}
\label{appdx:topic-modeling-human-justifications}

In the main paper, we examine the reasons driving the human's annotations of various perceived readability levels. 
To this end, we employ various automatic pattern extraction techniques, including frequency-based analysis (represented in the form of wordclouds) and n-gram feature importance. 
The following sections provide additional details about each of these experiments.

\subsection{Frequency-based Analysis}
\label{appdx:topic-modeling:word-cloud}

As part of our analysis, we conduct a frequency-based analysis of the rationales behind the readability judgments provided by the human annotators in the \eliwhygpt dataset. 

\paragraph{Methodology.} 
We conduct our analysis by first separating the dataset into three subsets according to the perceived readability level of the GPT4-generated explanations. 
In doing so, we obtain a total of 324, 694, and 182 examples corresponding to the \esshort, \hsshort, and \phdshort, respectively. 
Subsequently, we merge the annotators justification field for each subset, remove the English stopwords (as provided by the \textsc{nltk} library). 
To aggregate words with similar meanings, we further lemmatize each word using the \textsc{WordNetLemmatizer}\footnote{\url{https://www.nltk.org/api/nltk.stem.WordNetLemmatizer.html}}.


\begin{figure*}[tb]
  \centering
  \begin{subfigure}{0.32\linewidth}
    \includegraphics[width=\linewidth]{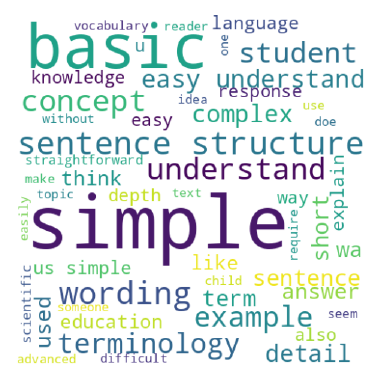}
    \caption{\es}
    \label{sfig:wordclouds-elementary}
  \end{subfigure}\hfill
  \begin{subfigure}{0.32\linewidth}
    \includegraphics[width=\linewidth]{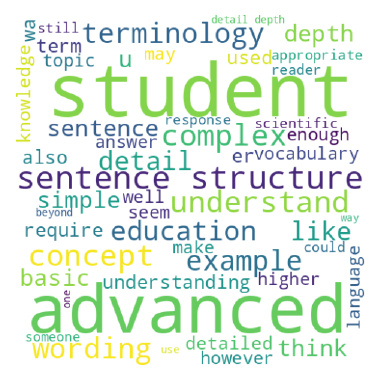}
    \caption{\hs}
    \label{sfig:wordclouds-highschool}
  \end{subfigure}\hfill
  \begin{subfigure}{0.32\linewidth}
    \includegraphics[width=\linewidth]{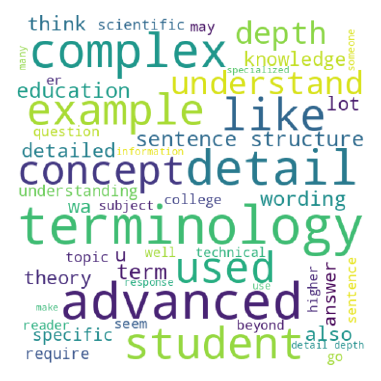}
    \caption{\phd}
    \label{sfig:wordclouds-graduate}
  \end{subfigure}
  \caption{Frequency-based analysis of the language expressions used by  human annotators when judging the perceived readability of various GPT4-generated explanations in \eliwhygpt. These word clouds are collected over 324, 694, and 182 examples annotated for \esshort, \hsshort, and \phdshort, respectively.}
  \label{fig:topic-modeling:wordclouds}
\end{figure*}

\subsection{Predictive Analysis}
\label{appdx:topic-modeling:logistic-regression}

We also conduct a model-based approach to determine the discriminative power of different phrases for each readability class. 
In this analysis, each annotator's justification is considered to be an individual document and both term and document frequencies are used to determine the readability class of a annotators' justifications. 

\paragraph{Methodology.} 
Similarly to the frequency-based analysis, we first decompose the \eliwhygpt dataset into three exclusive subsets based on the human perceived readability label. 
Additionally, we expand the justification field into individual documents, resulting in 707, 1665, and 416 total documents for \esshort, \hsshort, and \phdshort, respectively.
As preprocessing steps, we remove the English stopwords using the \textsc{nltk} default list, lemmatize the text using the \textsc{WordNetLemmatizer}, and lowercase the text.
Finally, we compute the term-to-document frequency matrix using \textsc{sklearn}'s \textsc{TfidfVectorizer}. 
To ensure that we capture complex phrases and not just individual words, we consider n-grams where $n \in \{1, 2, 3, 4\}$ and, to avoid overfitting to terms that appear in a single document, set \textsc{min\_df}=2. 

Having the term-to-document frequency matrix, we adopt a one-vs-all approach, where we iteratively fit a linear model to discriminate justifications of one class (\eg \esshort) from justifications outside of this class (\eg \hsshort and \phdshort). While focusing on linear models such as logistic regression allow us to directly examine the predictive importance of different n-grams, it pre-assumes that the most is a strong predictor. With the intent of selecting a good predictive model, we perform hyperparameter optimization using 10-fold cross-validation while using predictive accuracy as the evaluation criteria. 
We consider the following hyperparameters and employ grid search:
\begin{itemize}
\item \verb|estimator = LogisticRegression()|
\item \verb|max_iter = {100, 300}|
\item \verb|C = {0.01, 0.1, 1, 10, 100, 500}|
\item \verb|penalty = {l1, l2, elasticnet}|
\item \verb|solver = {liblinear, saga}|
\end{itemize}

We list the best obtained models for each readability class in Table~\ref{tab:appdx:predictive-power:hyperparam-configs}. 
Across all readability classes, we find that the fitted logistic regression outperforms a simple baseline that predicts the majority class (\textsc{Majority Accuracy}) by at least 3\% and up to 15\% absolute points. 

\begin{table*}[tb]
\centering
\caption{Hyperparameter configurations of the Logistic Regression models fit for each readability class. We use a grid search to find the optimal combination over the hyperparameters \textsc{C}, \textsc{penalty}, and \textsc{solver}. The best configuration is defined as the best achieving accuracy determined using 10-fold cross-validation.} 
\label{tab:appdx:predictive-power:hyperparam-configs}
\begin{tabularx}{0.9\linewidth}{l X c c}
\toprule
\textbf{Readability Class} & \textbf{Hyperparameters} & \textbf{Majority Accuracy (\%)} & \textbf{Best Accuracy (\%)} \\
\midrule
\esshort & 
\makecell[l]{
  C = 100 \\ 
  max\_iter = 300 \\
  penalty = l1 \\ 
  solver  = saga
} & 74.64 & 88.05 \\
\addlinespace
\hsshort & 
\makecell[l]{
    C = 500 \\
    max\_iter = 100 \\
    penalty = l1 \\
    solver = saga
} & 59.72 & 75.11 \\

\addlinespace
\phdshort & 
\makecell[l]{
  C = 100 \\
  max\_iter = 300 \\
  penalty = l1 \\ 
  solver  = saga
} & 85.08 & 88.34 \\

\bottomrule
\end{tabularx}
\end{table*}

\begin{figure*}[tb]
  \centering
  \begin{subfigure}{0.34\linewidth}
    \includegraphics[width=\linewidth]{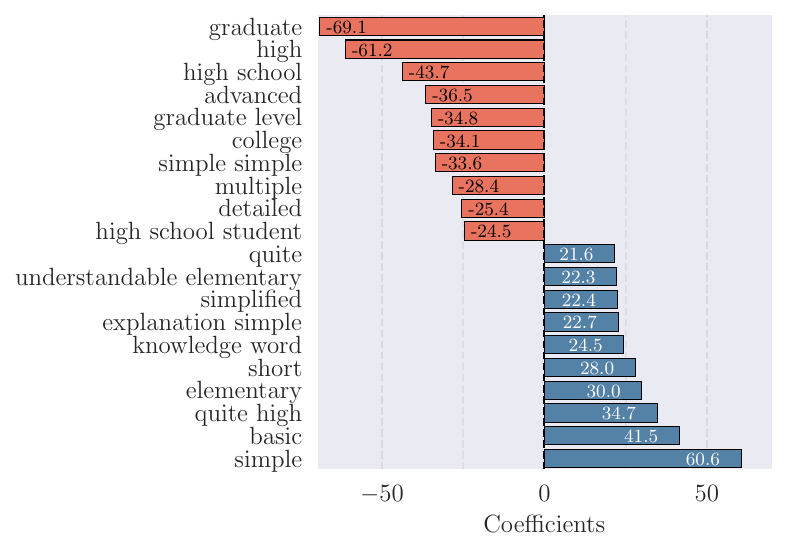}
    \caption{\es}
    \label{sfig:topic-modeling:top-10-features-elementary}
  \end{subfigure}\hfill
  \begin{subfigure}{0.34\linewidth}
    \includegraphics[width=\linewidth]{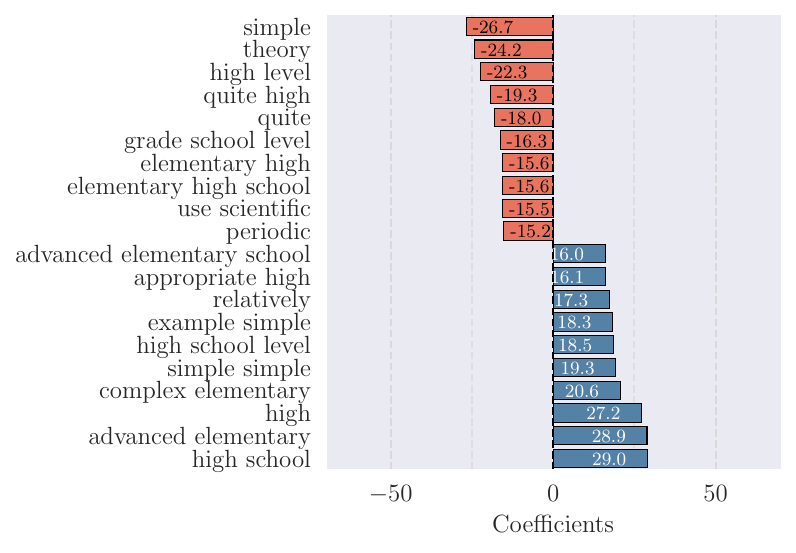}
    \caption{\hs}
    \label{sfig:topic-modeling:top-10-features-highschool}
  \end{subfigure}\hfill
  \begin{subfigure}{0.30\linewidth}
    \includegraphics[width=\linewidth]{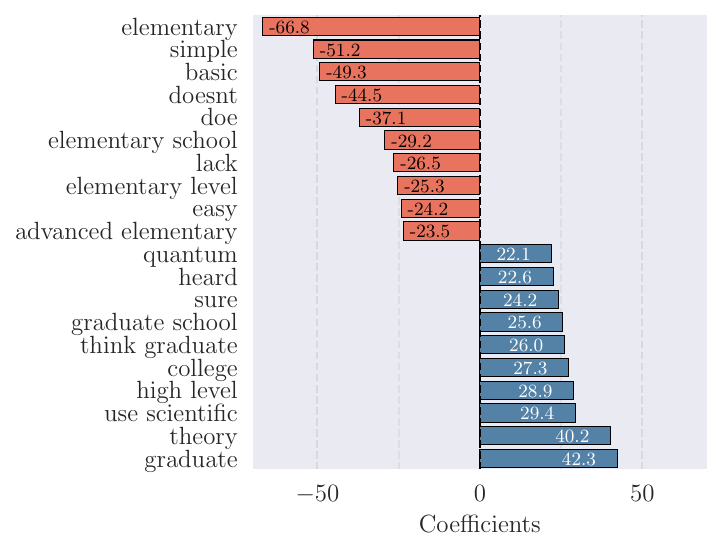}
    \caption{\phd}
    \label{sfig:topic-modeling:top-10-features-graduate}
  \end{subfigure}
  \caption{Regression analysis of the language expressions used by human annotators when judging the perceived readability of various GPT4-generated explanations in \eliwhygpt. These values clouds are collected over 324, 694, and 182 examples annotated for \esshort, \hsshort, and \phdshort, respectively.}
  \label{fig:topic-modeling:top-10-features}
\end{figure*}

\begin{table*}[tb]
\centering
\caption{Human rationales underlying readability judgments across 3 different readability classes: \esshort, \hsshort, \phdshort. Each row refers to the analysis of the same ``Why'' question but different GPT-4 explanation, being sourced from \eliwhygpt~\citep{joshi2025eliwhyevaluatingpedagogicalutility}.}
\label{tab:qualitative-examples:human-rationales}
\small
\begin{tabular}{p{0.30\linewidth} | p{0.30\linewidth} | p{0.30\linewidth}}
\toprule
\textbf{\esshort} & \textbf{\hsshort} & \textbf{\phdshort} \\
\midrule
\rowcolor{gray!10} 
    - It's probably too verbose for elementary levels, but I think people reading at that level could understand this explanation. The words are short enough. \newline
    - The explanation uses basic English language to interpret why humans are inclined towards social interactions. There are not many technical or professional terminologies, making it easy to understand. The sentence structures are simple, making it easy to follow.
    & 
    - Pretty easy and straight forward to understand. Not using complex words or scientific words. \newline
    - The sentences are short in length and easy to digest. It uses terms like ``elements'' and ``conductivity and ductility'' which require deeper understanding of elements and reactions. \newline
    - The explanation is written in a way that is easy to understand, but the details and some of the words used such as ``corrosion'' would make it difficult for an elementary reader to comprehend. However, the material is not so specialized that you would learn it on the graduate level, meaning this falls into the high school reading level.
    & 
    - The terminology seems higher level and more complicated than elementary or high school; \newline 
    - This is borderline HS/GS to me. But the terms ``parasocial'' and ``existential fears'' are a bit much for a typical high school student. It should be simplified a bit for an HS student.
    \\
\addlinespace 
    - The details are very surface level and it uses simple wording.
    - Simple sentence structure with simple  and short explanations. Not detailed or in depth. \newline
    - They used simple wording and examples to make their point.
    - It uses simple words like electricity, and can be easily understood \newline
    - It gives clear examples like copper being easy to stretch and not rusting, the sentences are short and straightforward. It gives enough detail to understand why copper is used in wires.
    &
    - The wording/terminology, examples, and details suggest high school-level engagement. It lacks the technicality of graduate school while being too advanced for elementary school; \newline
    - Using terminology like ``ritual'', ``theological'' and ``philosophical'' which requires basic knowledge of these terms. Depth and detail are also moderate levels but not quite a graduate level understanding; \newline
    - Wording Terminology, Sentence Structure, Details and depth
    & 
    - No way most high school students could follow this; \newline
    - The details and depth show of a graduate school person answering this.\\
\addlinespace 
\rowcolor{gray!10} 
     - Simple wording, a concept that most students of elementary school age should be able to grasp. Also not too many details. \newline
    - The explanation uses simple and direct language without complex terminology, making it accessible to children or adults with basic education. \newline
    - I think this text's wording, examples, sentence structure, and amount of detail are simple enough for an elementary-age student to comprehend. 
    & 
    - This response includes references to Alzheimer's, which I think would be outside the understanding of a typical 4th grader. It also references brain waves, which I think is covered in high school-level science courses.\newline
    - It uses more elevated vocabulary than Elementary School, however the lack of citations and more complex concepts and narrative structure make it less than Graduate School.
    & 
    - The language is more advanced and mentions more specific scientific theories. \newline
    - The amount of detail and specific terminology make me think it is a graduate level. \\
\bottomrule
\end{tabular}
\end{table*}

\section{Related Work}

In this section, we extend the discussion of readability metrics provided in the main paper. Specifically, we elaborate on the limitations of the previously proposed LLM-as-a-judge approaches and remaining challenges. 

\paragraph{Readability Assessment using LLMs.}
\citet{rooein-etal-2024-beyond} show that combining yes/no prompts with conventional metrics yields stronger correlations with human judgments than using either set of metrics alone. 
\citet{trott-riviere-2024-measuring} use 0-shot prompts to extract continuous readability scores which correlate strongly with human judgments.
In spite of promising results, these approaches have seen little adoption in practice.
Their reliance on repeated prompting introduces significant inference overhead, making them costly for large-scale evaluation or use as reward functions.
They also require allocating part of the already limited readability data to calibrate combinations or thresholds, further limiting their practicality.
Finally, although prior work has explored continuous readability assessments with LMs, to our knowledge their ability to distinguish coarse-grained readability classes remains unexplored.
\end{document}